\documentclass[10pt,twocolumn,letterpaper]{article}

\usepackage{booktabs}       %
\usepackage{color, colortbl}
\usepackage{adjustbox}
\definecolor{Gray}{gray}{0.9}

\usepackage{amsmath,amsfonts,bm}

\def\eqref#1{equation~\ref{#1}}

\def\1{\bm{1}}

\DeclareMathAlphabet{\mathsfit}{\encodingdefault}{\sfdefault}{m}{sl}
\SetMathAlphabet{\mathsfit}{bold}{\encodingdefault}{\sfdefault}{bx}{n}

\usepackage{iccv}
\usepackage{times}
\usepackage{epsfig}
\usepackage{graphicx}
\usepackage{amsmath}
\usepackage{amssymb}
\usepackage{url}
\usepackage{float}
\usepackage{theorem}
\usepackage{makecell}
\usepackage{amssymb}
\usepackage{pifont}
\usepackage{multirow}
\usepackage{multicol}
\usepackage{subcaption}
\usepackage{enumitem}
\usepackage{wrapfig}
\usepackage{adjustbox}
\usepackage{balance}

\newcommand{\crossvit}{CrossViT}
\newcommand{\deit}{DeiT}
\newcommand{\deitt}{DeiT-T}
\newcommand{\deittgraft}{DeiT-T+GrafT}

\newcommand{\vit}{ViT}
\newcommand{\swin}{Swin}
\newcommand{\swint}{Swin-T}
\newcommand{\swintgraft}{Swin-T+GrafT}

\newcommand{\cswin}{CSWin}
\newcommand{\cswinxt}{CSWin-XT}
\newcommand{\cswinxtgraft}{CSWin-XT+GrafT}
\newcommand{\cswint}{CSWin-T}
\newcommand{\cswintgraft}{CSWin-T+GrafT}
\newcommand{\ours}{GrafT}
\newcommand{\localmsa}{L-MSA}
\newcommand{\mobn}{MobileNet}
\newcommand{\mobv}{MobileViT}
\newcommand{\mobvxxs}{MobileViT-XXS}
\newcommand{\mobvxxsgraft}{MobileViT-XXS+GrafT}
\newcommand{\mobvxs}{MobileViT-XS}
\newcommand{\mobvxsgraft}{MobileViT-XS+GrafT}
\newcommand{\mobvs}{MobileViT-S}
\newcommand{\mobvsgraft}{MobileViT-S+GrafT}
\newcommand{\mobvv}{MobileViTv2}
\newcommand{\mobvva}{MobileViTv2-0.5}
\newcommand{\mobvvagraft}{MobileViTv2-0.5+GrafT}
\newcommand{\mobvvb}{MobileViTv2-1.0}
\newcommand{\mobvvbgraft}{MobileViTv2-1.0+GrafT}
\newcommand{\coco}{COCO 2017}
\newcommand{\imgdata}{ImageNet-1K}

\newcommand{\ccg}{\cellcolor{Gray}}

\def\eg{\emph{e.g.,}~} 
\def\ie{\emph{i.e.,}~}

\definecolor{azure(colorwheel)}{rgb}{0.0, 0.5, 1.0}

\newcommand{\cmark}{\ding{51}}
\newcommand{\xmark}{\ding{55}}

\newcommand{\PreserveBackslash}[1]{\let\temp=\\#1\let\\=\temp}
\newcolumntype{C}[1]{>{\PreserveBackslash\centering}p{#1}}
\newcolumntype{R}[1]{>{\PreserveBackslash\raggedleft}p{#1}}
\newcolumntype{L}[1]{>{\PreserveBackslash\raggedright}p{#1}}

\DeclareMathSymbol{\ast}{\mathbin}{symbols}{"03}

\usepackage[pagebackref=true,breaklinks=true,letterpaper=true,colorlinks,bookmarks=false]{hyperref}

\iccvfinalcopy %

\ificcvfinal\fi

\begin{document}

\title{Grafting Vision Transformers}

\author{\leftline{Jongwoo Park$^{1}$, Kumara Kahatapitiya$^{1}$, Donghyun Kim$^{2}$, Shivchander Sudalairaj$^{2}$,} 
\vspace{2mm}
\hspace{-1mm}
\\
\noindent{\leftline{Quanfu Fan$^{3}$\thanks{Part of work was done while at MIT-IBM AI Watson Lab}, Michael S. Ryoo$^{1}$}} \\ \\
\leftline{$^{1}$Stony Brook University}\\
\leftline{$^{2}$MIT-IBM Watson AI Lab} \\
\leftline{$^{3}$Amazon} \\ \\
\leftline{{\tt\small \{jongwopark@, kkahatapitiy@, mryoo@\}.cs.stonybrook.edu}} \\
\leftline{{\tt\small \{dkim, shiv.sr\}@ibm.com}} \\
\leftline{{\tt\small \{quanfu\}@amazon.com}} \\
}

\maketitle
\ificcvfinal\thispagestyle{empty}\fi

\begin{figure*}[t]
    \begin{subfigure}{.6\textwidth}
      \centering
      \hspace{15mm}
      \includegraphics[width=0.9\linewidth]{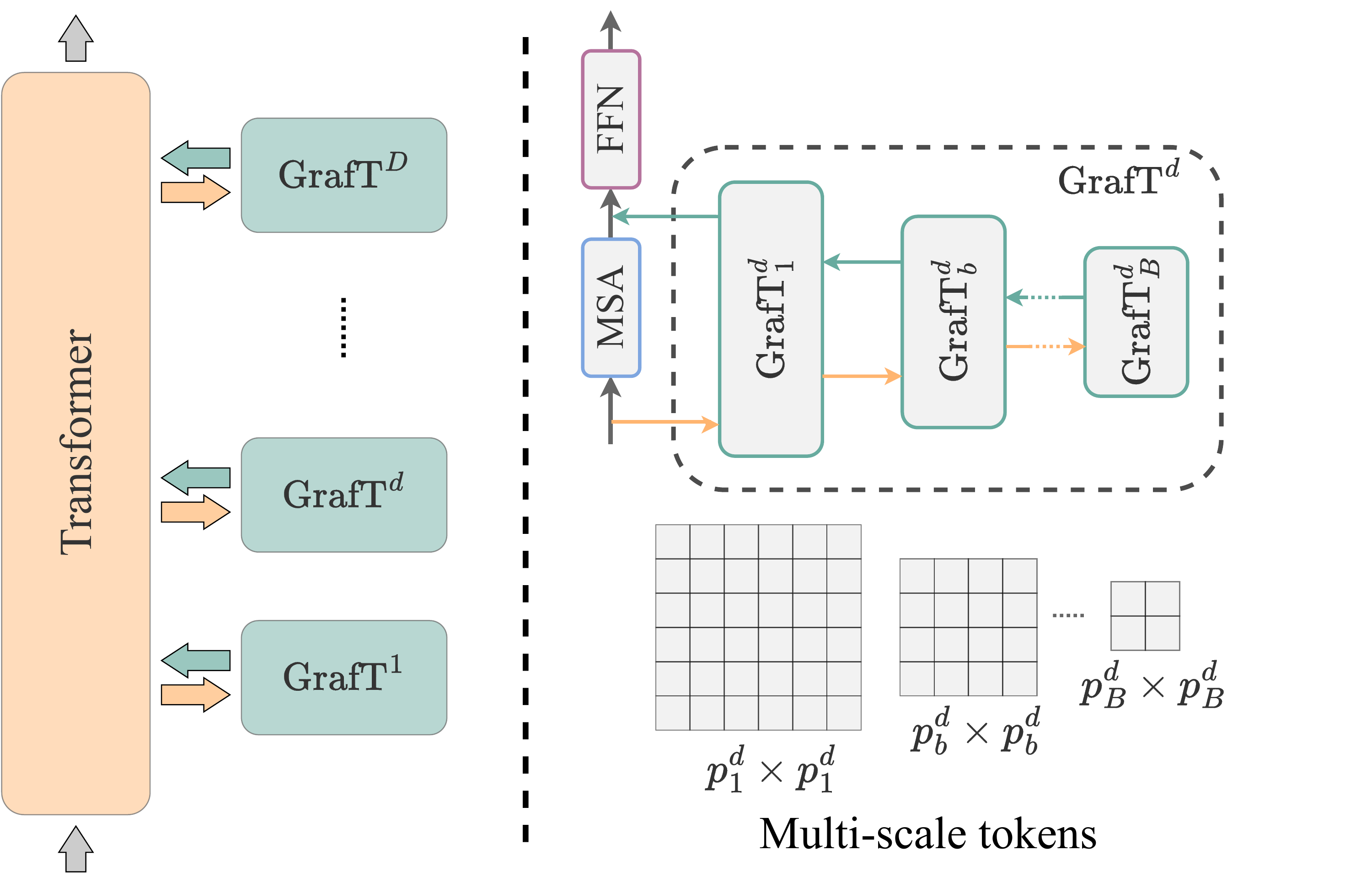}
      \vspace{-2mm}
      \caption{}  
      \label{fig:overview:left}
    \end{subfigure}
    \begin{subfigure}{.4\textwidth}
      \centering
      \hspace{-15mm}
      \includegraphics[width=0.9\linewidth]{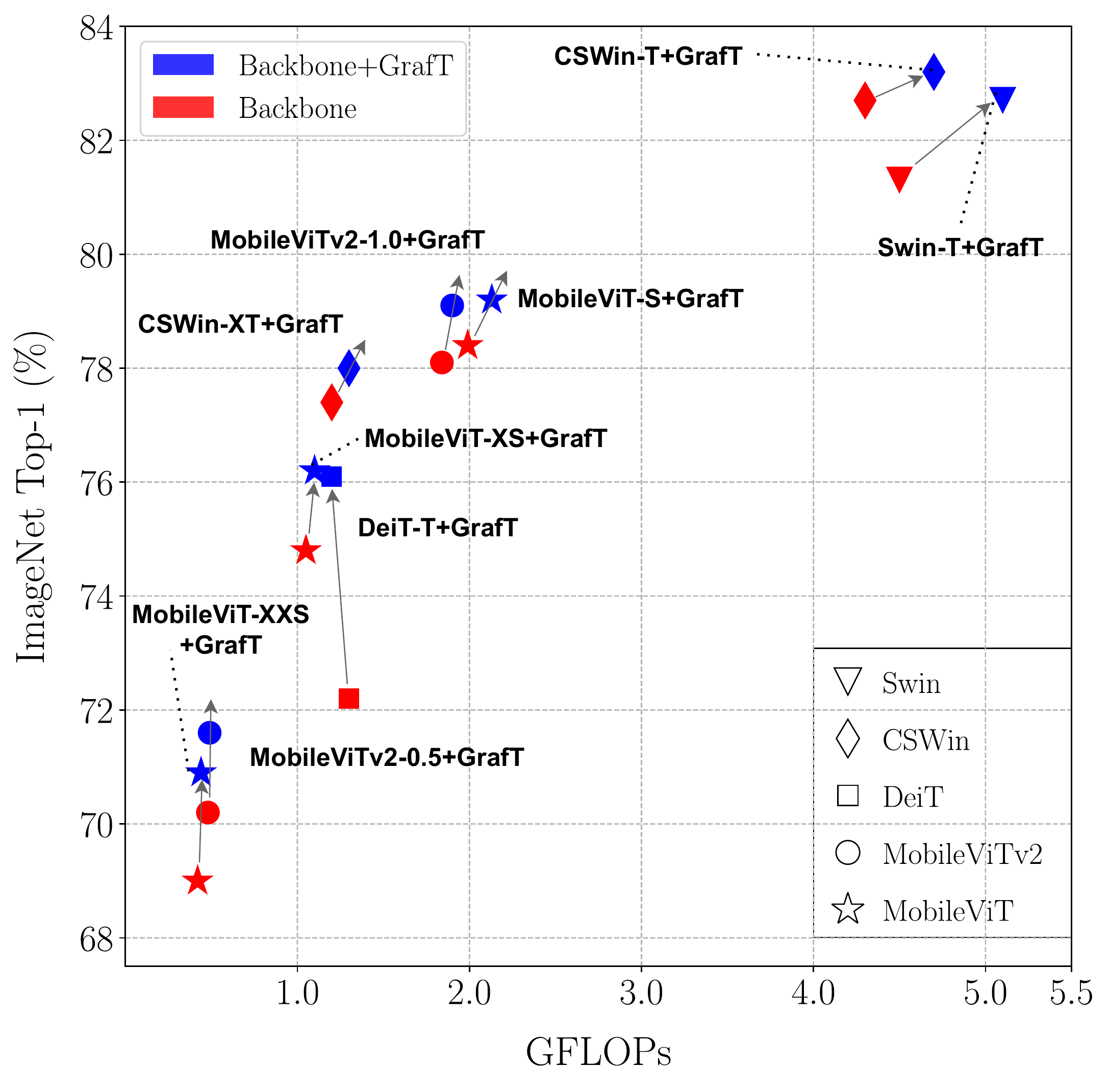}  
      \vspace{-2mm}
      \caption{}
      \label{fig:overview:right}
    \end{subfigure}
    \vspace{-6mm}
\caption{\small We introduce \ours, an add-on component that makes use of global and multi-scale dependencies at arbitrary depths of a network. (a) An overview of how GrafT modules are branched-out (or grafted) from a backbone Transformer and generate  multiple scales of features. (b) Performance-complexity trade-off of integrating GrafT in various backbones incorporating hybrid or pure architectures, homogeneous or pyramid structures, and various self-attention methods. \ours~shows consistent and considerable gains with a minimal increment in complexity. \deittgraft~uses fewer FLOPs because local attention is used to adopt GrafT as explained in \ref{ex:homo_pyramid}}
\label{fig:overview}
\end{figure*}
\begin{abstract}
   Vision Transformers (ViTs) have recently become the state-of-the-art across many computer vision tasks. In contrast to convolutional networks (CNNs), ViTs enable global information sharing even within shallow layers of a network, i.e., among high-resolution features. However, this perk was later overlooked with the success of pyramid architectures such as \swin~Transformer, which show better performance-complexity trade-offs. In this paper, we present a simple and efficient add-on component (termed \textbf{\ours}) that considers global dependencies and multi-scale information throughout the network, in both high- and low-resolution features alike. It has the flexibility of branching out at arbitrary depths and shares most of the parameters and computations of the backbone. GrafT shows consistent gains over various well-known models which includes both hybrid and pure Transformer types, both homogeneous and pyramid structures, and various self-attention methods. In particular, it largely benefits mobile-size models by providing high-level semantics. On the ImageNet-1k dataset, GrafT delivers \textbf{+3.9}\%, \textbf{+1.4}\%, and \textbf{+1.9}\% top-1 accuracy improvement to \deitt, \swint, and \mobvxxs, respectively. Our code and models will be made available.
\end{abstract}

\section{Introduction}
Self-attention mechanism in Transformers~\cite{NLP_transformer_bello2019attention} has been widely-adopted in language domain for some time now. It can look into pairwise correlations between input sequences, learning long-range dependencies. More recently, following the seminal work in Vision Transformers (ViT)~\cite{vit_16x16word}, the vision community has also started exploiting this property, showing state-of-the-art results on various tasks including classification, segmentation and detection, outperforming convolutional networks (CNNs)~\cite{he2016deep_residual_learning_resnet, tan2019efficientnet, howard2017mobilenets, sandler2018mobilenetv2, mobilenetv3_Howard_2019_ICCV, high_performance_nfnet_brock_2021}.
Motivated by this success, many variants of vision Transformers (\eg DeiT~\cite{deit_touvron21a}, \crossvit~\cite{crossvit_Chen_2021_ICCV}, TNT~\cite{TNT_NEURIPS2021}) emerged, inheriting the same homogeneous structure of \vit~(i.e., a structure w/o downsampling). 
However, due to the quadratic complexity of attention, such a structure becomes expensive, especially for high-resolution inputs and does not benefit from the semantically-rich information present in multi-scale representations.

To address these shortcomings, Transformers with pyramid structures (i.e., structures w/ downsampling) such as \swin~\cite{swin_Liu_2021_ICCV} were introduced with hierarchical downsampling and window-based attention, which can learn multi-scale representations at a computational complexity linear with input resolution. As a result, pyramid structures become more suited for tasks such as segmentation and detection. However, still, multiple scales arise deep into the network due to stage-wise downsampling, meaning that only the latter stages of the model may benefit from them. Thus, we poise the question: what if we can introduce multi-scale information even at the early stages of a Transformer without incurring a heavy computational burden? In particular, can CNN-based hybrid Transformers reap the benefits of GrafT as they heavily rely on the local information from CNNs?

Previous work has also looked into the direction above, both in CNNs~\cite{szegedy2015going} and in Transformers~\cite{crossvit_Chen_2021_ICCV, chen2022regionvit}. However, models such as CrossViT requires carefully tuning the spatial ratio of two feature maps in two branches and RegionViT needs considerable modifications to handle multi-scale training. To mitigate these issues, in this paper, we propose a simple and efficient add-on component called \textbf{\ours}~(see Figure~\ref{fig:overview}-(a)). It can be easily adopted in existing hybrid or pure Transformers, homogeneous or pyramid architectures, and various self-attenion methods, enabling multi-scale features throughout a network (even in shallow layers) and showing consistent performance gains while being computationally lightweight. \ours~is applicable at any arbitrary layer of a network. 

\begin{figure*}[t]
    \centering
    \includegraphics[width=\linewidth]{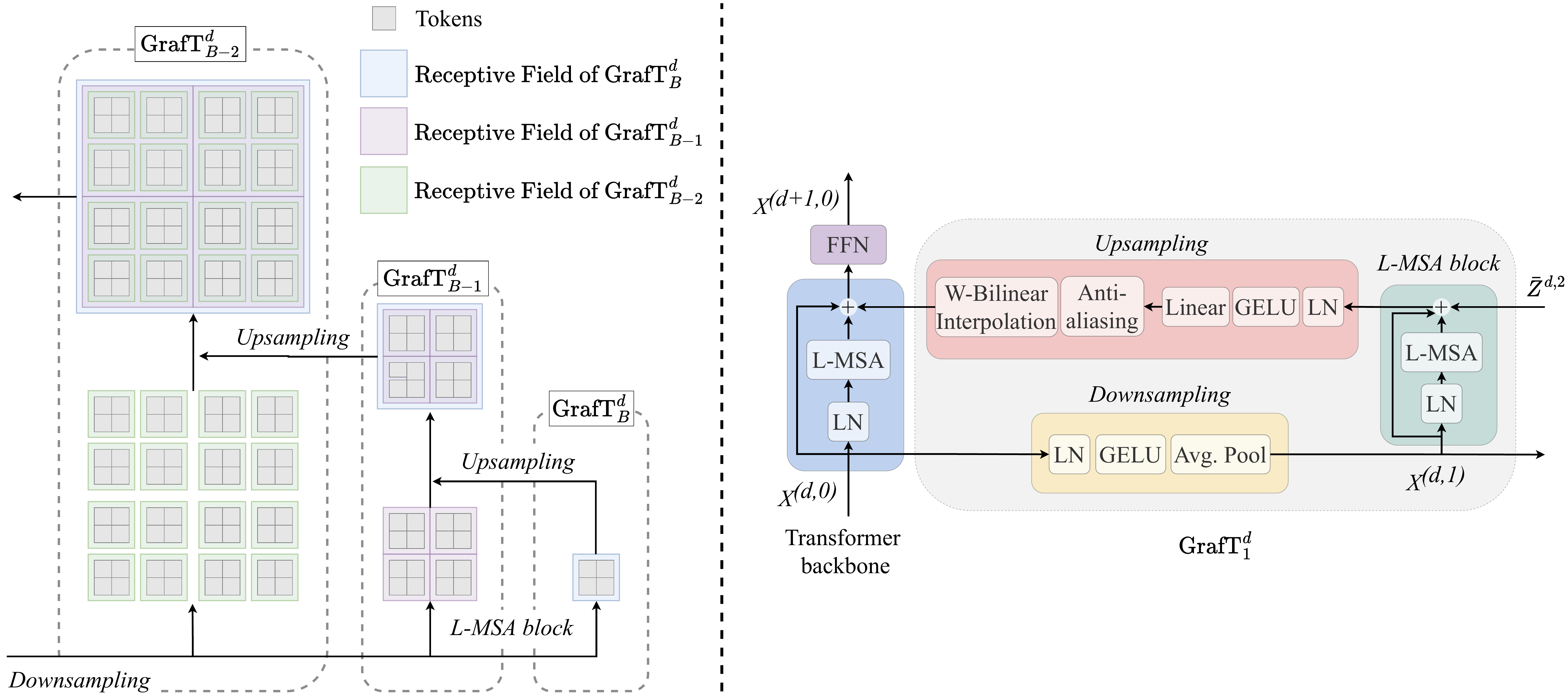}
    (a) \hspace*{9cm} (b) 
    \vspace{-2mm}
    \caption{\small An overview view of \ours. \textbf{(a):} Here, we show the receptive field of attention mechanisms in each scale and how it changes by merging with information from the corresponding lower-resolution scale (i.e., from branch-to-the-right). Multiple scale are created with downsampling, processed with local attention (in \textit{L-MSA})--- which becomes global attention in coarsest scale (\ours$^d_B$), and merged back with upsampling. When a lower-resolution representation is upsampled and merged, the effective receptive field increases, essentially giving access to efficiently-extracted global (or larger-local) information. \textbf{(b):} We present all the components and grafting/ merging points in \ours. We graft prior to self-attention block in the backbone, and merge prior to FFN so that we can reuse the heavy computations. Downsampling (\textit{left-right pathway}) uses light-weight average pooling to create a lower-resolution features, whereas upsampling (\textit{right-left pathway}) uses learnable window-based bi-linear interpolation (\textit{W-Bilinear}) to upscale. The processing unit within a \ours~module is a \textit{L-MSA} block, which performs local-attention. When merging features to higher-resolution, we use element-wise addition.}    
    \label{fig:fusion_window}
\end{figure*}

It consists of three main components: (1) a \textit{left-right pathway} for downsampling, (2) a \textit{right-left pathway} for upsampling, and (3) a \textit{bottom-up connection} for information sharing at each scale. The left-right pathway uses a series of average pooling operations to create a set of multi-scale representations. For instance, if \ours~ is attached to a layer with $(56\times 56)$ resolution, it can create scales of $(28\times 28)$, $(14\times 14)$ and $(7\times 7)$. We then process information at each scale with a \textit{\localmsa} block, a local self-attention mechanism (\eg window-attention)--- which becomes global-attention in the coarsest scale, as window-size becomes the same as the resolution. Next, the right-left pathway uses a series of learnable and window-based bi-linear interpolation (\textit{W-Bilinear}) operations to generate high-resolution features by upsampling the low-resolution outputs of \localmsa--- which contains global (or high-level) semantics extracted efficiently, at a lower resolution. Such upsampled features are merged with high-resolution features of the branch-to-the-left, which contain lower-level semantics, as also done in Feature Pyramid Networks~\cite{FPN_Lin_2017_CVPR}. Refer to Figure~\ref{fig:fusion_window}-(b) for a detailed view.

\ours~is unique in the sense that it can extract multi-scale information at any given layer of a Transformer while also being efficient. It relies on the backbone to do the heavy-lifting, by using a minimal computation overhead within grafted branches, in contrast to having completely-separate branches as in CrossViT~\cite{crossvit_Chen_2021_ICCV}.
In our evaluations, we observe that GrafT delivers gains to both light-weight hybrid (CNN + ViT) Transformers (\mobv~\cite{mehta2022mobilevit}, \mobvv~\cite{mehta2022separable}) and high-performing pure Transformers (DeiT~\cite{deit_touvron21a}, Swin~\cite{swin_Liu_2021_ICCV}, CSwin~\cite{cswin_Dong_2022_CVPR}). In particular, GrafT helps light-weight hybrid models become high-performing general-purpose networks with a minimal increase in complexities (Parameters and FLOPs). For the light-weight network \mobvxxs, GrafT increases the top-1 accuracy by \textbf{+1.9}\% in classification as shown in Table~\ref{table:pyramid_img1k_type} and improves the mAP by \textbf{+0.8}\% in object detection as shown in Table~\ref{table:det_ssd} while adding minimal complexities. In addition, GrafT can be smoothly integrated with various self-attention methods as it worked well with regular multi-head self-attention (MHSA) in DeiT, shifted-window self-attention in Swin, cross-shaped window self-attention in CSWin, inter-patch self-attention in \mobv, and separable self-attention in \mobvv. We summarize the performance of GrafT by grouping models into homogeneous (\vit~\cite{vit_16x16word}) and pyramid (\swin~\cite{swin_Liu_2021_ICCV} and \mobn~\cite{howard2017mobilenets}) architectures. On ImageNet-1K~\cite{deng2009imagenet}, GrafT improves the top-1 accuracy by \textbf{+3.9}\% for DeiT-T as shown in Table~\ref{table:homogeneous_img1k}, by \textbf{+1.9}\% for \mobvxxs~by \textbf{+1.4}\% for \swint, and by \textbf{+0.5}\% for \cswint ~as shown in Table~\ref{table:pyramid_img1k_type}. We also observe significant gains in performance when grafted models are used as feature backbones for object detection and segmentation tasks. We believe multi-scale high-level semantics from GrafT help models to identify objects in various sizes. On \coco~\cite{lin2014microsoft}, GrafT provides \textbf{+1.6} mAP for \mobvxs, \textbf{+1.1} AP$^\text{b}$ and \textbf{+0.8} AP$^\text{m}$ for Swin-T. On ADE20K~\cite{zhou2017scene} semantic segmentation, GrafT provides \textbf{+1.0} mIOU$^\text{ss}$, \textbf{+1.3} mIOU$^\text{ms}$ with \swin-T+\ours. Figure \ref{fig:overview}-(b) shows the performance-complexity trade-off when integrating GrafT in the backbone on ImageNet-1K.

\section{Grafting Vision Transformers}
\label{sec:method}
Our goal is to provide multi-scale global information to the backbone Transformer from the bottom layer so that high-level semantics from GrafT can help the Transformer to construct more efficient features. Since Graft is modular, it can be applied to various Transformer architectures. We select backbones incorporating various architectural characteristics to show that GrafT is a general-purpose module. The backbones cover hybrid and pure Transformer, homogeneous (ViT~\cite{vit_16x16word}) and pyramid (Swin~\cite{swin_Liu_2021_ICCV}, MobileNet~\cite{howard2017mobilenets}) structures, and different types of self-attention methods.

\subsection{Overall Architecture}
The overall architecture of GrafT is illustrated in Figure ~\ref{fig:fusion_window}-(a). We start with a backbone Transformer (\ie \vit~or \swin) and simply attach GrafT to some vertical layers of the backbone Transformer. For an input image with size of $H \times W \times 3$, the patch tokens to the first vertical layer is $\frac{H}{4} \times \frac{W}{4} \times C$ after the patch embedding. 

 \ours~is a horizontal pyramid structure which consists of left-right pathway (series of \ours~downsampling), right-left pathway (series of \ours~upsampling), and bottom-up connection (multiple \ours~\localmsa) as shown in Figure~\ref{fig:fusion_window}-(b). The input feature to \ours~goes through left-right pathway (downsampling), right-left pathway (upsampling), bottom-up connection (\localmsa) and becomes a feature having strong high-level semantics at multiple scales which then gets fused into the backbone Transformer. Specifically, for the GrafT attached to the vertical layer at $S^{th}$ stage, the input feature to the GrafT has the size of $\frac{H}{4r^{S-1}} \times \frac{W}{4r^{S-1}} \times C$ where $r=2$ for pyramid structure and $r=1$ for homogeneous structure. Then, we fuse the feature from \ours~to the original backbone, which will be described in the later section.
 
 \subsection{Transformer+GrafT Blocks}
 In this section, we describe each operation in Figure ~\ref{fig:fusion_window}-(b). Let $X^{d,b}$ as an input tensor at vertical layer $d$ and horizontal downsampling level $b$ in GrafT with the shape: $X^{d,b} \in \mathbb{R}^{H^b \times W^b \times C}$. $H^b$, $W^b$, $C$ is the height, width, channels of the feature map $X^{d,b}$ at a horizontal level $b$ respectively.

 \noindent\textbf{Left-right pathway (downsampling):}
 The left-right pathway creates serial feature maps at several scales with the downsampling rate r which follows the vertical pyramid downsampling rate. For example, in the first stage of Swin+GrafT, GrafT creates three downsampled feature maps \{$X^{1}, X^{2}, X^{3}$\} by downsampling the input feature map $X^{0}$ from the backbone with downsampling rate 2. In left-right pathway, we use adaptive average pooling for downsampling:

\begin{equation}
    \begin{split}
    \label{eq:ds}
        X^{d,b+1} = A(X^{d,b}) = \rho(\text{GELU}(\text{LN}(X^{d,b})))
    \end{split}
\end{equation}

  $A$ is a lightweight downsampling function which consists of LayerNorm ($LN$), $GELU$, and adaptive average pooling ($\rho$) , which is more cost-efficient than other downsampling methods such as cross attention and linear projection (see Table~\ref{table:abl_deit_ds_us}). It maps the input $X^{d,b}$ to downsampled feature $X^{d,b+1}$: $\mathbb{R}^{H^{b} \times W^{b}} \mapsto \mathbb{R}^{H^{b+1} \times W^{b+1}} $ where $H^{b+1} < H^{b}, W^{b+1} < W^{b}$. $X^{b+1}$ is the input at the horizontal downsampling level $b+1$. Downsampling happens sequentially over horizontal layers to progressively abstract the fine information in coarse features.
 
 \noindent\textbf{Right-left pathway (upsampling):}
 The right-left pathway hallucinate serial higher resolution feature maps by upsampling low resolution feature maps. These upsampled feature maps are enhanced by feature maps from bottom-up connection that retain spatially more accurate activations and lower-level semantics. This enhancement process is similar to FPN~\cite{FPN_Lin_2017_CVPR}. In right-left pathway, GrafT upsampling uses learnable W-Bilinear (window-base bilinear) interpolation which is more cost-efficient than other upsampling methods such as cross attention and nearest neighbor interpolation as shown in Table~\ref{table:abl_deit_ds_us}. Learnable W-Bilinear interpolation solves the aliasing problem by embedding anti-aliasing weights, the sigmoid of positional embeddings, in the feature maps. Anti-aliasing weights learns perturbations for each grid in the feature map that can prevent aliasing effect. W-Bilinear interpolation is adopted to address the semantics discontinuity between local regions. Finally, our upsampling is defined as:

\begin{align}
    \begin{split}
    \label{eq:us}
        \bar{Z}^{d,b+1} &= \Phi(\tilde{Z}^{d,b+1}) \\
        &= \Phi(E_{aa}\odot\alpha(Z^{d,b+1})) \\
        &= T(Z^{d,b+1}) 
    \end{split} 
        \\
        \bar{Z}^{d,b+1}_{u_m,v_n} &= \Phi({\tilde{Z}}^{d,b+1}_{i_m,j_n})
\end{align}

$T$ is a lightweight upsampling function mapping the representation back to the spatial resolution in the previous horizontal level. This function is designed to upsample output features from \localmsa. First, given the tensor $Z^{d,b+1}$, the output from \localmsa, channel mixing ($\alpha$) is applied to align channels before interpolation. $\alpha$ consists of LayerNorm ($LN$), $GELU$, and a linear layer. Next, anti-aliasing embeddings ($E_{aa}$) are multiplied to resolve the aliasing problem. The proposed anti-aliasing embeddings are the output of sigmoid function on position embeddings $Z^{j+1}_{pos} \in R^{H^{j+1} \times W^{j+1} \times C}$. It learns to provide perturbations in the spatial dimension that prevents interpolation from suffering the aliasing problem. It is a simpler and lighter method compared to 3x3 convolutions~\cite{FPN_Lin_2017_CVPR}. Lastly, W-Bilinear interpolation ($\Phi$) maps $\tilde{Z}^{d,b+1} \in \mathbb{R}^{H^{b+1} \times W^{b+1}}$ to $\bar{Z}^{d,b+1} \in \mathbb{R}^{H^{b} \times W^{b}}$. It can be described as mapping the low resolution feature in each $(m, \, n)$th window  $\tilde{Z}^{d,b+1}_{i_m,j_n}$ into high resolution feature in $(m, \, n)$th window $\bar{Z}^{d,b+1}_{u_m,v_n}$. $(i_{m}, \, j_{n})$ is a spatial position $(i, \, j)$ in $(m, \, n)$th window where $i \in I, \, j \in J, \, m \in M, \, n \in N$. $(u_{m}, \, v_{n})$ is a spatial position $(u, \, v)$ in $(m, \, n)$th window where $u \in r_{h}I, v \in r_{w}J$. $r_{h}, \, r_{w}$ are the height and width ratio between feature resolutions in two consecutive horizontal levels. Upsampling happens sequentially over horizontal layers to progressively upscale spatial resolution of coarse features.

\noindent\textbf{\localmsa~in GrafT:}
The local self-attention method from the backbone Transformer, limits the self-attention in a local region which causes the discrepancy of semantics at the boundary of local regions. Therefore, bilinear interpolation is applied in each local region multiple times instead of the entire feature map at once. The bottom-up connection merges lower-level feature maps enhanced by \localmsa~and higher-level feature maps upsampled by W-Bilinear interpolation by element-wise addition. The merging process iterates until it generates the high-level feature map that has the same spatial size as the input feature map $X^{0}$ from the backbone. In the example above, the coarsest feature map $X^3$ goes through \localmsa~and becomes $\bar{Z}^{3}$, a feature map having the highest-level semantics. $\bar{Z}^{2}$ is generated by merging upsampled $\bar{Z}^{3}$ with output feature from \localmsa~which applies local self-attention on $X^2$. This process iterates until the finest feature map $\bar{Z}^0$ which has the same spatial size of the $X^0$ is produced. $\bar{Z}^0$ is then fused into the output feature map from \localmsa~in the backbone Transformer and proceed to FFN block. 

\begin{equation}
    \begin{split}
    \label{eq:w_msa}
        Z^{d,b+1} = X^{d,b+1} + [\text{\localmsa}(\text{LN}(X^{d,b+1})) + \bar{Z}^{d,b+2}]
    \end{split}
\end{equation}

It is a simple and light block that fuses multi-scale features. It uses a standard window-based MSA from Swin~\cite{swin_Liu_2021_ICCV} to encode fine features $X^{d,b+1}$. It uses simple element-wise addition to fuse this fine feature and coarse feature $\bar{Z}^{d,b+2} \in \mathbb{R}^{H^{b+1} \times W^{b+1}}  $coming from one deeper horizontal level. Lastly, Skip connection is added to produce output $Z^{d,b+1} \in \mathbb{R}^{H^{b+1} \times W^{b+1}} $

\noindent\textbf{Bottom-up connection with multiple high-level semantics:}
In GrafT, \localmsa~uses local self-attention to enhance downsampled feature maps but it limits receptive field within each local region. It is important to exchange information among local regions so that feature can embed not only local structure but also global structure. In the left side of Figure~\ref{fig:fusion_window}, the features at the bottom are the output of \ours~\localmsa~where receptive field is limited to each local region. The feature map at $b+2$ level $\tilde{Z}^{b+2}$ has a receptive field, a blue color line, that covers the entire feature map. The feature map at $b+1$ level, $\tilde{Z}^{b+1}$, originally suffers the information discrepancy between local regions due to local receptive field. When $\tilde{Z}^{b+2}$ is merged into $\tilde{Z}^{b+1}$, $\tilde{Z}^{b+1}$ inherits a receptive field from $\tilde{Z}^{b+2}$ and this newly added global receptive field provides higher-level semantics that can understand the relation between local regions. $\tilde{Z}^{b}$, the feature map at $b$ level, also originally suffers the information discrepancy between local regions. When $\tilde{Z}^{b+1}$ is merged into $\tilde{Z}^{b}$, $\tilde{Z}^{b}$ inherits receptive fields from both $\tilde{Z}^{b+1} \: \tilde{Z}^{b+2}$ and this newly added receptive fields provides multi-scale high-level semantics that can understand the relation between local regions in multiple aspects. Multi-scale receptive fields which are progressively generated from multi-scale features resolve the drawback of local receptive field formed by \localmsa~in the backbone Transformer. The \localmsa in the original backbone Transformer is defined as:

\begin{equation}
    \begin{split}
    \label{eq:tf_msa}
        Y^{d,0} = X^{d,0} + [\text{\localmsa}(X^{d,0}) + \bar{Z}^{d,1}]
    \end{split}
\end{equation}

\localmsa~is the local multi-self attention that the Transformer in the main branch is using to encode fine feature $X^{d,0} \in \mathbb{R}^{H^0 \times W^0 \times C}$. The encoded fine feature is element-wise added with coarse feature $\bar{Z}^{d,1}$ from the GrafT. Lastly, skip connection is added to produce output $Y^{d,0} \in \mathbb{R}^{H^0 \times W^0 \times C}$. It is interesting that a simple element-wise addition successfully fuse the fine feature in the main branch and the coarse feature from GrafT. Thanks to the power of GrafT upsampling method for the robust fusion of the multi-scale features encoded by various MSA. Then, we use the same FFN in the backbone transfomers:

\begin{equation}
    \begin{split}
    \label{eq:tf_ffn}
        X^{d+1,0} = Y^{d,0} + \text{MLP}(\text{LN}((Y^{d,0})))
    \end{split}
\end{equation}

It is a standard Transformer block performing channel mixing via LayerNorm ($LN$) and $MLP$ on the output of \localmsa~block and adds the skip connection to generate output $X^{d+1,0}$ which is the input to the next vertical layer.

\noindent\textbf{Computation complexity:} 
\label{sec:method:comp_complexity}
Average pooling and bilinear interpolation runs in $\Theta(HW)$ and \localmsa~in GrafT follows the complexity of \localmsa~in the backbone Transformer. Since \localmsa~is more complex than $\Theta(HW)$, the complexity of Transformer+GrafT is equal to the complexity of the pure backbone Transformer. For example, 
\begin{equation}
    \begin{split}
    \centering
    \Omega{(\text{Swin})} = \Omega{(\text{Swin+GrafT})} = {12HWC^2 + 2M^2HWC}
    \end{split}
\end{equation}
 where H,W is the width and height of feature map and M is the size of window.
 
\section{Experiments}
\subsection{ImageNet-1K Classification}
\noindent{We integrate GrafT in various models to show that it is generally applicable. We observe consistent gains by attaching GrafT on models that are hybrid or pure Transformers, have homogeneous or pyramid structures, and exploit various self-attention methods. We train our models on the \imgdata~with the standard settings in \cite{vit_16x16word, swin_Liu_2021_ICCV, cswin_Dong_2022_CVPR, cswin_Dong_2022_CVPR, mehta2022mobilevit, mehta2022separable}. Additional details are written in the appendix.
\begin{table}[t]
\centering
\caption{\small Performance of GrafT with homogeneous architectures on ImageNet-1K \cite{deng2009imagenet}. \deittgraft~outperforms \deitt~\cite{deit_touvron21a} by +3.9\% and even surpasses PVT-T \cite{PVT_Wang_2021_ICCV} (a pyramid structure).}
\vspace{-2mm}
\label{table:homogeneous_img1k}
\resizebox{1.0\columnwidth}{!}{
\begin{tabular}{l|crrr}
\toprule
\multirow{2}{*}{Model} & \multirow{2}{*}{Structure} & Params $\downarrow$  & FLOPs $\downarrow$ & Acc. $\uparrow$ \\
& & (M) & (G) & (\%)\\
\midrule
\deitt~\cite{deit_touvron21a} & Homoge. & 5.7 & 1.3 & 72.2 \\
\rowcolor{Gray}
\deittgraft & Homoge. &  7.9 & 1.2 & \textbf{(+3.9) 76.1} \\
CrossViT-9~\cite{crossvit_Chen_2021_ICCV} & Homoge. & 8.6 & 1.8 & 73.9 \\
PVT-T~\cite{PVT_Wang_2021_ICCV} & Pyramid & 13.2 & 1.9 & 75.1 \\
\bottomrule
\end{tabular}
}
\vspace{-2mm}
\end{table}
\begin{table}[t]
\centering
\caption{\small Performance of GrafT with pyramid architectures on ImageNet-1K \cite{deng2009imagenet}. Graft shows consistent gains across various architectures (Pure/Hybrid), model sizes and attention mechanisms.}
\label{table:pyramid_img1k_type}
\vspace{-2mm}
\resizebox{1.0\columnwidth}{!}{
\begin{tabular}{L{36mm}|C{12mm}rR{13mm}R{17mm}}
\toprule

\multirow{2}{*}{Model} & \multirow{2}{*}{Type} & Params $\downarrow$  & FLOPs $\downarrow$ & Acc. $\uparrow$ \\
& & (M) & (G) & (\%)\\
\midrule
\mobvxxs \cite{mehta2022mobilevit} & Hybrid & 1.27 & 0.42 & 69.0 \\
\rowcolor{Gray}
\mobvxxsgraft & Hybrid & 1.43 & 0.44 & (+1.9) 70.9 \\
\mobvva \cite{mehta2022separable} & Hybrid & 1.37 &  0.48 & 70.2 \\
\rowcolor{Gray}
\mobvvagraft & Hybrid & 1.53 & 0.49 & \textbf{(+1.4) 71.6} \\
\midrule
MobileFormer-52 \cite{chen2021mobileformer} & Hybrid & 3.5 & 52M & 68.7 \\
\mobvxs \cite{mehta2022mobilevit} & Hybrid & 2.32 & 1.05 & 74.8 \\
\rowcolor{Gray}
\mobvxsgraft & Hybrid & 2.65 & 1.10 & \textbf{(+1.4) 76.2} \\
\midrule
\mobvs \cite{mehta2022mobilevit} & Hybrid & 5.58 & 1.99 & 78.4 \\
\rowcolor{Gray}
\mobvsgraft & Hybrid & 6.38 & 2.13 & \textbf{(+0.8) 79.2} \\
\mobvvb \cite{mehta2022separable} & Hybrid & 4.90 &  1.84 & 78.1 \\
\rowcolor{Gray}
\mobvvbgraft & Hybrid & 5.52 & 1.90 & (+1.0) 79.1 \\
ViL-Tiny-RPB~\cite{ViL_Zhang_2021_ICCV} & Transformer & 7 & 1.3 & 76.7 \\
\cswinxt* \cite{cswin_Dong_2022_CVPR} & Transformer & 6 & 1.2 & 77.4 \\
\rowcolor{Gray}
\cswinxtgraft & Transformer & 8 & 1.3 & (+0.6) 78.0 \\

\midrule
PVT-M~\cite{PVT_Wang_2021_ICCV} & Transformer &   44 & 6.7 & 81.2 \\
PoolFormer-S36~\cite{poolformer_Yu_2022_CVPR} & Transformer &   31 & 5.2 & 81.4 \\
$\text{T2T}_{t}$-14~\cite{T2T_Yuan_2021_ICCV} & Transformer & 22 & 6.1 & 81.7 \\
TNT-S~\cite{TNT_NEURIPS2021} & Transformer & 24 & 5.2 & 81.5 \\
ViL-S-RPB~\cite{ViL_Zhang_2021_ICCV} & Transformer &   25 & 4.9 & 82.4 \\
RegionViT-S~\cite{chen2022regionvit} & Transformer &   31 & 5.3 & 82.6 \\
\swint~\cite{swin_Liu_2021_ICCV} & Transformer &   29 & 4.5 & 81.3 \\
\rowcolor{Gray}
\swintgraft & Transformer & 34 & 5.1 & (+1.4) 82.7 \\
\cswint \cite{cswin_Dong_2022_CVPR} & Transformer & 23 & 4.3 & 82.7 \\
\rowcolor{Gray}
\cswintgraft & Transformer & 29 & 4.7 & \textbf{(+0.5) 83.2} \\
\bottomrule
\end{tabular}
}
\vspace{-2mm}
\end{table}

\label{ex:homo_pyramid}

\noindent\textbf{Homogeneous or pyramid structures: } Table~\ref{table:homogeneous_img1k} shows the models that has homogeneous structures (i.e., with a constant spatial resolution, or \#tokens) . DeiT-T+GrafT achieves +3.9\% boost over DeiT-T. Even though DeiT underperforms CrossViT (-1.7\%), DeiT-T+GrafT outperforms it by +2.2\% as it receives high-level semantics from GrafT. Interestingly, even though DeiT-T+GrafT does not take advantage of the vertical pyramid structure, it even outperforms PVT (which is a vertical pyramid structure).

When implementing DeiT-T+GrafT, we replace global attention in the backbone with window-based local attention (w/o window shifting) to separate local-global processing as in our motivation (refer Figure~\ref{fig:fusion_window}-(b)). We confirm that this slightly-modified DeiT (prior to applying GrafT) underperforms the original DeiT by 2\%. Thus, overall, \ours~delivers +5.9\% performance gain (from 70.2\% to 76.1\%).

Table~\ref{table:pyramid_img1k_type} contains models that adopt a  pyramid structure. GrafT consistently boosts performance in \mobv (+1.9\% in -XXS, +1.4\% in -XS, +0.8\% in -S), \mobvv (+1.4\% in -v0.5, +1.0\% in -v1.0), \swin (+1.4\% in -T), and \cswin (+0.6\% in -XT$^*$, +0.5\% in -T) with minimal increase in complexities. Models with SOTA accuracies at each scale are highlighted in bold. These results show that GrafT generalizes well.

\noindent\textbf{Hybrid or pure Transformer:}
One research direction follows pure Transformers, whereas another follows hybrid models (i.e., CNNs+Transformers). GrafT performs well in both. When applied in pure Transformers (\deit, \swin, \cswin), it shows consistent gains (eg: +1.4\% in \swintgraft). When applied in powerful, light-weight hybrid models (\mobv, \mobvv), it also shows better performance (eg:  +1.9\% \mobvxxsgraft).

\noindent\textbf{Various self-attention methods:}
As shown in Figure \ref{fig:fusion_window}-(b), GrafT inherits the self-attention operation from the backbone that it is applied in. Therefore, we explore whether GrafT works with different self-attention mechanisms. For instance, in \deitt~(w/ global MSA), it obtains +3.9\%, whereas in \swint~(w/ shifted-window local MSA) it obtains +1.4\%. \cswint~utilizes cross-shaped MSA, in which GrafT shows +0.5\%. In \mobvxxs~(w/ inter-path MSA) and \mobvva~(w/ separable MSA), GrafT gives +1.9\% and +1.4\% boosts, respectively. We observe that GrafT provides consistent gains by inheriting the specific type of self-attention in each backbones.

\subsection{Object Detection and Segmentation}

\noindent{We benchmark GrafT on object detection and segmentation, showing its capabilities as a general-purpose model. We consider various backbones such as pyramid structures, hybrid architectures an pure Transformers, each having different self-attention mechanisms.}

\begin{table}[t]
    \centering
    \caption{\small Performance of \ours~with mobile backbones on a single shot object detection task on the \coco~\cite{lin2014microsoft}. GrafT consistently improves the detection performance of \mobv~\cite{mehta2022mobilevit}.}
    \vspace{-2mm}
    \resizebox{1.0\columnwidth}{!}{
    \begin{tabular}{l|crrr}
    \toprule
    \multirow{2}{*}{Model} & \multirow{2}{*}{Type} & Params $\downarrow$  & FLOPs $\downarrow$ & Acc. $\uparrow$ \\
    & & (M) & (G) & (\%)\\
    \midrule
         \mobvxxs\cite{mehta2022mobilevit} & Hybrid & 1.7 & 0.90 & 19.9 \\ 
    \rowcolor{Gray}
         \mobvxxsgraft  & Hybrid & 1.9 & 0.91 & \textbf{(+0.7) 20.6} \\ 
    \midrule
         MobileNetv1 \cite{howard2017mobilenets} & CNN & 5.1 & 1.3 & 22.2 \\
         MobileNetv2 \cite{sandler2018mobilenetv2} & CNN & 4.3 & 0.8 & 22.1 \\
         MobileNetv3 \cite{howard2019mobilenetv3} & CNN & 5.0 & 0.6 & 22.0 \\
         MobileViT-XS \cite{mehta2022mobilevit} & Hybrid & 2.7 & 1.89 & 24.8 \\ 
    \rowcolor{Gray}
         MobileViT-XS+GrafT  & Hybrid & 3.1 & 1.98 & \textbf{(+1.6) 26.4} \\
    \midrule
         MobileViT-S \cite{mehta2022mobilevit} & Hybrid & 5.7 & 3.48 & 27.7  \\
    \rowcolor{Gray}
         MobileViT-S+GrafT  & Hybrid & 6.5 & 3.65 & \textbf{(+1.1) 28.8}  \\
    \bottomrule
    \end{tabular}
    }
    \vspace{-2mm}
    \label{table:det_ssd}
\end{table} 
\begin{table}[t]
    \centering
    \caption{\small Performance of \ours~on two-stage object detection and instance segmentation on \coco~\cite{lin2014microsoft}. GrafT outperforms Swin\cite{swin_Liu_2021_ICCV}. Here, $1\times$ (SS) corresponds to 12 epochs with single scale, $3\times$ (MS) corresponds to 36 epochs with multi-scale.}
    \vspace{-2mm}
    \label{table:object_detection}
    \resizebox{1.0\columnwidth}{!}{
    \begin{tabular}{l|R{10mm}R{10mm}|R{10mm}R{10mm}|R{10mm}R{10mm}}
    \toprule
    \multirow{2}{*}{Model} & \multirow{2}{*}{Params$\downarrow$} & \multirow{2}{*}{FLOPs$\downarrow$} & \multicolumn{2}{c|}{1x (SS)} & \multicolumn{2}{c}{3x (MS)} \\
                           &                             &                            & \multicolumn{1}{c}{AP$^\text{b}$$\uparrow$}     & \multicolumn{1}{c|}{AP$^\text{m}$$\uparrow$}     & \multicolumn{1}{c}{AP$^\text{b}$$\uparrow$}     & \multicolumn{1}{c}{AP$^\text{m}$$\uparrow$}     \\
    \midrule
    PVT-S~\cite{PVT_Wang_2021_ICCV}                   & 44                          & 245                        & 40.4        & 37.8        & 43.0        & 39.9        \\
    Swin-T~\cite{swin_Liu_2021_ICCV}                  & 48                          & 264                        & 42.2        & 39.1        & 46.0        & 41.6        \\
    \rowcolor{Gray} & & & 43.3 & 39.9 & 47.0 & 42.5 \\
    \rowcolor{Gray} \multirow{-2}{*}{Swin-T+GrafT} & \multirow{-2}{*}{53} & \multirow{-2}{*}{275}  & \textbf{(+1.1)} & \textbf{(+0.8)} & \textbf{(+1.0)} & \textbf{(+0.9)} \\
    RegionViT-S~\cite{chen2022regionvit}             & 50                          & 171                        & 42.5        & 39.5        & 46.3        & 42.3        \\
    VIL-S-RPB~\cite{ViL_Zhang_2021_ICCV}                   & 45                          & 277                        & --      & --        & 47.1        & 42.7        \\
    \cswint~\cite{cswin_Dong_2022_CVPR} & 42 & 279 & 46.7 & 42.2 & 49.0 & 43.6 \\

    \bottomrule
    \end{tabular}
    }
\end{table}

\begin{figure*}[t]
    \centering
    \includegraphics[width=1.0\linewidth]{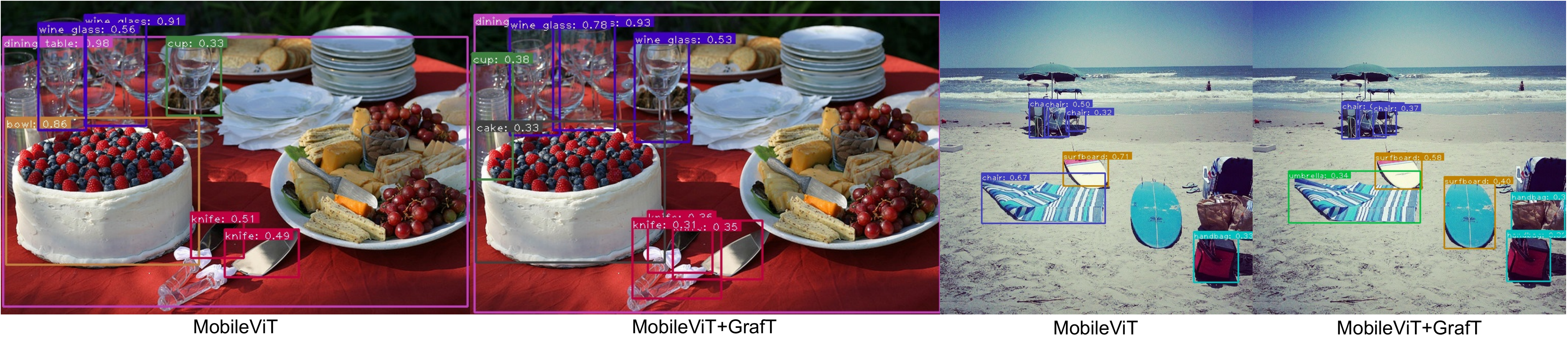}
    \vspace{-4mm}
    \caption{\small{Object detection results of \mobvxs~\cite{mehta2022mobilevit} and~\mobvxsgraft~on the \coco~\cite{lin2014microsoft} validation set. In the left figure, GrafT correctly detects cake, cup, and wine glass correctly in contrast to the baseline. In the right figures, GrafT correctly detects surfboard and handbag. It shows to be capturing multi-scale information better.}}
    \label{fig:detection_vis}
    \vspace{-2mm}
\end{figure*}
\noindent\textbf{Experiments on object detection and instance segmentation:} We run single shot and two-stage object detection on the \coco~\cite{lin2014microsoft} by following a standard settings in \mobv~\cite{liu2016ssd, mehta2022mobilevit} and \swin~\cite{swin_Liu_2021_ICCV}. The detailed settings are in the appendix. In Table~\ref{table:det_ssd}, we show the benefit of GrafT in single-shot object detection. In \mobvxs, it gains +1.6\% mAP while only increasing parameters by 12\% and FLOPs by 5\%. It even outperforms MobileNetv3 by 4.4\% with a 60\% model-size. GrafT helps the mobile-size Transformers to detect various sizes of objects by providing multi-scale features as shown in Fig~\ref{fig:detection_vis}. The results of \mobvv~are placed in the supplementary.

Table~\ref{table:object_detection} presents two-stage object detection and instance segmentation settings. \swintgraft~shows consistent gains in both $1\times$ and $3\times$ schedules. Even though Swin-T underperforms RegionViT-S, our Swin-T+GrafT outperforms it, showing an overall gain of +1.1 AP$^\text{b}$ +0.8 AP$^\text{m}$ in 1x(SS), +1.0 AP$^\text{b}$, +0.9 AP$^\text{m}$ in 3x(MS). This implies that GrafT is applicable as a general-purpose model.

\begin{table}[t]
\centering
    \caption{\small Performance of \ours~on semantic segmentation on ADE20K \cite{zhou2017scene}. GrafT outperforms Swin \cite{swin_Liu_2021_ICCV}.}
    \vspace{-2mm}
    \label{table:semantic_segmentation}
    \resizebox{1.0\columnwidth}{!}{
    \begin{tabular}{l|rrrr}
    \toprule
    \multirow{2}{*}{Model} & Params $\downarrow$  & FLOPs $\downarrow$ & mIOU $\uparrow$ & mIOU $\uparrow$ \\
    & (M) & (G) & (SS) & (MS)\\
    \midrule
    Swin-T~\cite{swin_Liu_2021_ICCV} & 59 & 945 & 44.5 & 45.8 \\
    \rowcolor{Gray}
    Swin-T + Graft & 66 & 955 & (+1.0) 45.5 & (+1.3) 47.1 \\
    \bottomrule 
    \end{tabular}
    }
    \vspace{-2mm}
\end{table}

\noindent\textbf{Experiments on semantic segmentation:}
For semantic segmentation, we train Swin-T on ADE20K~\cite{zhou2017scene} while following a
standard training procedure similar to Swin~\cite{swin_Liu_2021_ICCV} We provide implementation details in the appendix. Table~\ref{table:semantic_segmentation} shows that Swin+GrafT outperform Swin in both single-scale (by +1.0\%) and multi-scale (by +1.3\%) settings. The multi-level semantics from GrafT has enabled our model to capture better pixel-level details.

\subsection{Ablations on ImageNet-1K}
We conduct the following experiments to better understand the design decisions of GrafT. Here, we consider Swin-T and Deit-T backbones.

\begin{table}[t]
\caption{\small Different upsampling (US) and downsampling (DS) approaches considered in GrafT, evaluated on ImageNet-1K \cite{deng2009imagenet}. We fix the upsampling method when ablating downsampling methods, and vice-versa. Learn. W-Bilinear interpolation and avg. pooling are the best methods.}
\label{table:abl_deit_ds_us}
    \centering
    \resizebox{1.0\columnwidth}{!}{
    \begin{tabular}{c|c|rrrr}
        \toprule
        Fixed & Varied & Params $\downarrow$ & FLOPs $\downarrow$ & Thr. $\uparrow$ & Acc. $\uparrow$ \\
        sampling & sampling & (M) & (G) & (im/s) & (\%) \\
        \midrule
        \multirow{3}{*}{Linear proj.} & Nearest nei. (US) & 8.4 & 1.3 & 2890 & 74.8 \\
        & Cross att. (US) & 8.7 & 1.3 & 2392 & 74.4 \\
        & \ccg Learn. bilin. (US)  & \ccg 8.5 & \ccg 1.3 & \ccg 2260 & \ccg 75.2 \\
        \midrule
        \multirow{3}{*}{Learned bilin.} & Linear proj. (DS) & 8.5 & 1.3 & 2260 & 75.2 \\
        & Cross att. (DS) & 7.9 & 1.2 & 2834 & 75.4 \\
        & \ccg Avg. pool. (DS) & \ccg 7.9 & \ccg 1.2 & \ccg 3143 & \ccg 76.1 \\
        \bottomrule
    \end{tabular}
    }
\end{table}

\noindent\textbf{Downsampling \& upsampling in GrafT:.} Table~\ref{table:abl_deit_ds_us} shows the performance of DeiT-T+GrafT with different horizontal downsampling and upsampling approaches. We fix downsampling to be a linear projection when ablating upsampling approaches. Note that, in Cross attention, the finer-level features (in backbone) are used as as query while coarser-level features (in GrafT) acts as key/value. We also explore Nearest neighbor interpolation as the simplest upsampling method. Learnable Bilinear (i.e., Window-based Bilinear) interpolation uses anti-aliasing weights, and is applied in each local region separately. The results show that Learnable W-Bilinear interpolation achieves the highest accuracy with a reasonable complexity and speed.

We fix upsampling to be Learnable W-Bilinear when ablating downsampling approaches. Linear projection basically concatenates neighboring tokens and applies a linear layer (similar to Patch Merging in Swin). Cross attention first creates a coarser-level feature by average pooling a fine feature in the backbone (as query), using the backbone feature as key/value to perform cross attention. Average pooling simply creates a coarser feature by pooling. The results show that the simple average pooling achieves a better trade-off. Therefore, we adopt learnable W-Bilinear interpolation as upsampling and average pooling as downsampling in GrafT, by default.

\noindent\textbf{\#scales in GrafT:} In GrafT, we explore a horizontal pyramid structure where multi-scale low-resolution features are created. Here, we try to understand whether such multiple high-level semantics are useful, going beyond a single-scale. In Swin, we consider a maximum of $8\times$ horizontal downsampling (in 3 scales), similar to its original vertical downsampling. Table~\ref{table:abl_multi-scale_glbtk} shows that higher the \#scales, the better. Swin-T+GrafT with 3-scales gives +1.4\% with a little overhead compared to single-scale. Here, at the first stage of Swin (where inpiut resolution is $56 \times 56$), GrafT creates additional features of ($28 \times 28$), ($14 \times 14$), ($7 \times 7$) resolutions to deliver multi-scale global information, even at the start of the network.

\begin{table}[tb]
    \centering
    \caption{\small {Considering (a) the different number of scales within a GrafT at each layer, and (b) the different number of GrafTs along vertical layers, evaluated on ImageNet-1K \cite{deng2009imagenet}.}}
    \vspace{-3mm}
    \begin{subtable}[h]{0.45\textwidth}
        \caption{\small {}}
        \label{table:sota_21k}
        \centering
\label{table:abl_multi-scale_glbtk}
\begin{adjustbox}{max width=\linewidth}
    \begin{tabular}{l|c|rrr}
        \toprule
         \multirow{2}{*}{Model} & \multirow{2}{*}{\#scales} & Params $\downarrow$ & FLOPs $\downarrow$ & Acc. $\uparrow$ \\
          & & (M) & (G) & (\%)\\
        \midrule
        Swin-T~\cite{swin_Liu_2021_ICCV} & 0 & 29 & 4.5 & 81.3 \\
        \midrule
        \multirow{3}{*}{Swin-T + GrafT}
        & 1 & 33.5 & 5.0 &  (+1.0) 82.3 \\
        & 2 & 33.9 & 5.1 & (+1.2) 82.5 \\
        & \ccg 3 & \ccg 34.0 & \ccg 5.1 & \ccg \textbf{(+1.4) 82.7} \\
        \bottomrule 
    \end{tabular}
\end{adjustbox}
    \end{subtable}
    \begin{subtable}[h]{0.45\textwidth}
        \caption{\small }
        \vspace{-1.5mm}
        \centering
\label{table:abl_n_graft}
 \begin{adjustbox}{max width=\linewidth}
    \begin{tabular}{l|c|rrr}
        \toprule
         \multirow{2}{*}{Model} & \multirow{2}{*}{\#grafts} & Params $\downarrow$ & FLOPs $\downarrow$ & Acc. $\uparrow$ \\
          & & (M) & (G) & (\%)\\
         \midrule
         DeiT-T~\cite{deit_touvron21a} & 0 & 5.7 &	1.3 & 72.2 \\
         \midrule
         & 4 & 6.5 & 1.2 &  (+2.2) 74.4\\
         & 8 & 7.3 & 1.2 &  (+3.4) 75.6\\
         \multirow{-3}{*}{DeiT-T + GrafT}
         & \ccg 11 & \ccg 7.9 & \ccg 1.2 & \ccg \textbf{(+3.9) 76.1}\\
         \bottomrule
    \end{tabular}
    \vspace{-2mm}
\end{adjustbox}
    \end{subtable}
    \label{table:ablations_1}
    \vspace{-2mm}
\end{table}

\noindent\textbf{\#grafts in GrafT:} Table~\ref{table:abl_multi-scale_glbtk} shows that the accuracy consistently increases with \#grafts in DeiT. Therefore, we attach GrafT to all vertical layers after the first layer ($11\times$). It is not applicable in the first layer, because the input needs to be encoded first before being used to create coarser features through GrafT.

\begin{table}[t]
    \centering
    \caption{\small{Sharing the parameters of backbone-FFN by training the DeiT-T on ImageNet-1K}}
    \label{table:share_ffn}
    \vspace{-2mm}
    \resizebox{0.8\columnwidth}{!}{
        \begin{tabular}{c|ccr}
            \toprule
             Shared FFN & \multicolumn{1}{c}{Params (M) $\downarrow$} & \multicolumn{1}{c}{FLOPs (G) $\downarrow$} & \multicolumn{1}{c}{Acc. (\%) $\uparrow$}   \\
            \midrule
             \xmark & 8.2 & 1.2 & 74.8 \\
            \midrule
            \cmark & 7.9 & 1.2 &  \textbf{(+1.3) 76.1} \\
            \bottomrule
        \end{tabular}
    }
    \vspace{-2mm}
\end{table}

\noindent\textbf{Sharing FFN when grafting:} Table~\ref{table:share_ffn} compares the trade-offs of having either a shared FFN or two separate FFNs, using a DeiT-T backbone. When having separate FFNs, GrafT features are fused after FFNs, and when having a shared FFN, fusion happens before it. We see that a shared FFN achieves +1.3\% higher accuracy with fewer parameters compared to its counterpart. Therefore, we adopt a shared FFN design in the GrafT by default.

\section{Related Work}

\noindent\textbf{Vision Transformers:}
Convolution neural networks (CNNs) have been widely adopted as it have shown promising performance ~\cite{krizhevsky2012imagenet_cnns, he2016deep_residual_learning_resnet, chen2017dual_path_network, howard2017mobilenets, sandler2018mobilenetv2, hu2018squeeze_excitation, huang2017densely, simonyan2014very_deep, szegedy2015going_deeper_cnn, tan2019efficientnet} on small-scale dataset such as ImageNet-1K~\cite{deng2009imagenet}. Inductive biases such as translation invariance and locality from CNNs are the key reasons to be trained well from scratch in small-scale dataset. Recently, Transformers (\eg ViT~\cite{vit_16x16word} or DeiT~\cite{deit_touvron21a}) achieved comparable results to CNNs. The first type is a pure Transformer with a homogeneous structure like ViT where the number of tokens and channels do not change over the vertical layers. T2T~\cite{T2T_Yuan_2021_ICCV} proposes a progressive tokenization method where spatial structures are preserved. CrossViT~\cite{crossvit_Chen_2021_ICCV} creates two branches to formulate both local and global information and exchange information. The second type is a pure Transformer with a pyramid structure such as PiT~\cite{PiT_Heo_2021_ICCV} and PVT~\cite{PVT_Wang_2021_ICCV} where vertical layers are divided into multiple stages, and the number of tokens
is progressively decreased while the channel size increases over stages. Swin~\cite{swin_Liu_2021_ICCV} introduces shifted-window self-attention where self-attention is performed in each window and shifting window mechanism exchanges information among windows in a pyramid structure. RegionViT~\cite{chen2022regionvit} creates two branches to formulate local tokens and global tokens like CrossViT and assign each global token to local tokens in the same region to exchange information. CSWin~\cite{cswin_Dong_2022_CVPR} introduces cross-shaped window self-attention where half of the channels are used to create vertical stripes as local regions, and the other half is used to create horizontal stripes as local regions. The third type is a light-weight hybrid Transformer with a pyramid structure. Researchers have designed a hybrid Transformer where CNNs are combined with a Transformer~\cite{chen2021mobileformer, mehta2022mobilevit, mehta2022separable, li2022efficientformer} to compete with the well-studied light-weight CNNs~\cite{he2016deep_residual_learning_resnet, howard2017mobilenets, sandler2018mobilenetv2, howard2019mobilenetv3}. For example, \mobv~\cite{mehta2022mobilevit} places light-weight MobileNet blocks in the early stages to capture local features and exploits Transformer blocks in the late stages to capture global features. This process successfully incorporates spatial inductive biases through CNNs in Transformers. As a result, the model size is substantially reduced while stability and performance are improved.

In this paper, we propose GrafT, a simple and cost-efficient add-on that provides rich global information to backbone Transformers where there is a lack of communication between local regions because of local self-attention. 

\noindent\textbf{Exploiting multi-scale global tokens:} 
While pyramid structure Transformers (\eg Swin) learn multi-scale features in a hierarchical way, high-level semantics is introduced at the last layers and the Transformer cannot receive early guidance on how to efficiently encode low-level semantics by understanding the high-level semantics.

Some Transformers such as CrossViT~\cite{crossvit_Chen_2021_ICCV}, TNT~\cite{TNT_NEURIPS2021}, RegionViT~\cite{chen2022regionvit} keep two branches to encode low-level semantics and high-level semantics from the early stage. However, having two separate branches are detrimental to throughput and requires careful design of choosing which layers to exchange local and global information and choosing the right size ratio of low-resolution and high-resolution features as mentioned in CrossViT~\cite{crossvit_Chen_2021_ICCV} or the divisibility between local and global tokens. ViL~\cite{ViL_Zhang_2021_ICCV} creates global tokens through random initialization in each layer and use them to exchange information between local regions. However, this global token does not contain good inductive bias of local tokens and does not incorporate multiple high-level semantics. 
GrafT is unique in the sense that it delivers multiple high-level semantics by exploiting horizontal pyramid structure and uses simple element-wise addition to fuse global information to the backbone Transformer. It is applicable to Transformers with both homogeneous structure and pyramid structure and improves the performance of Transformers without increasing the computation complexity due to the light-weight components as described in \ref{sec:method:comp_complexity}. We provide additional related work and the differences from the prior work in the appendix.

\section{Conclusion}

In this paper, we introduced \ours: an add-on component that can easily be adopted in hybrid and pure Transformers, homogeneous and pyramid structures, and various self-attention methods, enabling multi-scale feature fusion in arbitrary depths of a model. 
The proposed \ours~branches are designed to be efficient, relying on the backbone to perform heavy computations. 
In fact, it gives consistent gains at a minimal computation burden. We also observe its effectiveness across multiple backbones and various benchmarks, including classification, detection, and segmentation. In the current work, GrafT is applied to three well-known Transformers: \deit, \swin, and \cswin~and two-well known hybrid Transformers: \mobv, and \mobvv. In particular, GrafT largely benefits mobile-size models because global information is crucial to understand the scenes at that accuracy level. Going forward, we hope that \ours~becomes a generally used component for introducing multi-scale features in Transformers.

{\small
\balance
\bibliographystyle{ieee_fullname}
\bibliography{egbib}
}

\end{document}